# Neural Information Organizing and Processing – Neural Machines

Iosif Iulian Petrila*

Faculty of Automatic Control and Computer Engineering, Gheorghe Asachi Technical University of Iasi, Str. Dimitrie Mangeron, Nr. 27, 700050, Iasi, Romania

The informational synthesis of neural structures, processes, parameters and characteristics that allow a unified description and modeling as neural machines of natural and artificial neural systems is presented. The general informational parameters as the global quantitative measure of the neural systems computing potential as absolute and relative neural power were proposed. Neural information organizing and processing follows the way in which nature manages neural information by developing functions, functionalities and circuits related to different internal or peripheral components and also to the whole system through a non-deterministic memorization, fragmentation and aggregation of afferent and efferent information, deep neural information processing representing multiple alternations of fragmentation and aggregation stages. The relevant neural characteristics were integrated into a neural machine type model that incorporates unitary also peripheral or interface components as the central ones. The proposed approach allows overcoming the technical constraints in artificial computational implementations of neural information processes and also provides a more relevant description of natural ones.

**Introduction**

In nature, neural information processing emerged with the necessity of complex beings such as animals to manage information in a more dynamic manner to ensure their interaction with an unpredictable environment, and implicitly their survival, by specialize certain cells for information management purpose through the nervous system as an integrative component of sensing and actuating (afferent and efferent) external and internal information by developing functions aimed to ensuring the adaptive and resilient characteristics. Through computing systems, the neural type of information processing began to be investigated for its perspective of use within artificial computing systems for understanding and validating natural neural processes through simulations as well for constructing devices with neural specific functions, consequently, various neural information aspects being highlighted over the time [1-17].

The neural information processing methods, although they have a tortuous history and do not have yet a coherent model not even at a fundamental level, are beginning to prove convincing in areas where specifically neural processes are more relevant than purely digital ones, such as: natural languages processing (text transformations, speech recognition and synthesis, etc.), patterns recognition, creative processing (multimedia), data analysis (healthcare, finance), control and coordination (robotics), etc. [7 - 17]. Since in essence the neural processes realize functional transformations of some contextual information, they are find usefulness in related fields such as formal systems, computing tools (compilers, libraries), etc. [18, 19]. Also, some common characteristics of neural phenomena with other ones open perspectives of combining them, offering possibility of better performing mutually modeling, such as with: genetic, fuzzy, quantum, etc. [20 - 22].

However, despite fragmented niche successes and enthusiastic perspectives, current neural methods present numerous limitations in artificial implementations compared to their natural counterparts, mostly caused by the lack of a unitary fundamental model to describe the neural specificity of information structuring and processing. The methods are deficient in continuous and plastic learning and adaptation due to the use of fixed topologies or architectures and training methods (such as backpropagation) suitable only for pre-training otherwise generate catastrophic forgetting on training with new data, necessitating retraining from scratch with all information old and new to preserve previous knowledge. The implementations of neural methods are computationally inefficient, requiring important resources for elementary neural operations compared to the natural case, the cause being inadequate algorithmic modeling and hardware support for distributed and parallel processes. Current models do not integrate the neural peripheral areas, being embodiment lacunars and difficult to adapt to real autonomous systems such as robotic ones because they do not integrate the neural management of sensors and actuators and are not scalable down to less powerful computing systems and with many categories of afferent and efferent components. The neural phenomena have been investigated from an informational perspective with focusing on central processes (see FNN, MLP, RNN, SVN, LSTM, CNN, GAN, etc.)[8 - 15] and neglecting the processing of peripheral or interface neural information, although in nature, neural processes were first developed around peripheral components and later incorporated into more complex processes in the form of afferent or efferent information by the central area that actually become an integrative extension of the peripheral areas. An aspect that created "winter" periods in the relevant modeling of neural information phenomena is the constraint or expectation of descriptions or processes to be deterministic from phenomena that have a non-deterministic foundation. Even if, from an application point of view, deterministic methods are the ones of interest, offering prediction and control, the deterministic constraints of neural models or algorithms remove their essence and only by adding some non-deterministic characteristics do they acquire characteristics similar to natural ones. Along with these, the linear modeling of some non-linear and dynamic phenomena as well as the symmetrization and homogenization of neural structures represented simplifications that, although easy from an implementation point of view, generated unrealistic results eventually overcome with significant costs.

---

* **E-mail address:** IosifIulianPetrila@gmail.com

**Neural Information**

Living nature has built its own genetic and neural information management tools in order to handle the interaction with a non-deterministic environment, and we are part of the category of beings that use such tools to ensure our survival in interaction with an unpredictable nature. The computer systems, in order to become more relevant instruments to human needs and to assist or replace certain related activities, must be upgraded with facilities that include processes as much as possible close to natural processes and especially neural ones. These processes must not be oversimplified to correspond to a technical level of the moment because they lose their relevance by generating informational catastrophes, and also must not follow biological structuring literally but only to preserve the essence of natural phenomena. Consequently, the basic informational characteristics description of neural systems must include information structures with real topologies (with asymmetries, feedbacks, heterogeneities, plasticity, etc.) and must not be limited to layer-type idealizations that are easy to approach in computing systems but that do not correspond to real neural entities and implicitly they will not realistically describe the related processes.

The informational quantitative potential of neural processes, as a scale measure, can be highlighted by defining specific parameters that can describe the computing power of neural entities, regardless of their specificity, natural or artificial. In this respect, if we analyzing the way in which the nature has organized the neural specificity of information processing through the specialization of some cells for this purpose and connecting them, it is observed that not the neural cells direct count confer processing power, but rather the capacity to fragment the information and make connections and correlations between these informational fragments through the connections of neural nodes (cells), biologically related to synaptic elements. Therefore, the absolute neural power can be defined as a measure or scale of the neural system capacity to correlate or connect information fragments through connections as:

$$\text{Neural Absolute Power} = \text{Log}_2 \text{Connections} \qquad (1)$$

in which the logarithm achieves a retention of connections power factor and base 2 allows a description compatible with binary information, obviously, in artificial systems the connections being managed by related parameters. In order to make a comparison with the human specific level of neural information processing, a relative neural power can also be introduced through:

$$\text{Neural Relative Power} = 100 \times \text{Log}_2 \text{Connections} / \text{Log}_2 \text{Connections}_{Human} \qquad (2)$$

in which the power of different neural systems can be represented as a percentage relative to the average human neural absolute power. Some representative natural and artificial neural entities with the related neural parameters can be seen in Table 1.

| Neural Entity | Neural Nodes | Neural Connections | Neural Absolute Power | Neural Relative Power |
|---|---|---|---|---|
| Ciona | 231 | 8617 | 13.07 | 27 |
| Honey Bee | $960 \times 10^3$ | $1 \times 10^9$ | 29.89 | 63 |
| African Elephant | $257 \times 10^9$ | $1.0 \times 10^{14}$ | 46.50 | 98 |
| Human | $86 \times 10^9$ | $1.5 \times 10^{14}$ | 47.09 | 100 |
| Hysteron | 1 | 2 | 1 | 2 |
| ChatGPT 3.5 | $> 0.5 \times 10^6$ | $175 \times 10^9$ | 37.34 | 79 |
| ChatGPT 4.0 | $> 1.3 \times 10^6$ | $1.76 \times 10^{12}$ | 40.67 | 86 |
| Material Universe | $10^{80}$ (Atoms) | $10^{160}$ | 531.50 | 1128 |

**Table 1 Neural Entities**

One can see the neural relevance of connections compared to that of neural nodes count on natural and also on artificial ones from the simplest one (the hysteron - elementary contextual memory unit with history) up to the present most complex one (ChatGPT). At the end, the entire material universe was added by extrapolation as a neural entity just to have an upper limit benchmark (taking into consideration only one type of interaction between all atoms, even if at long distances it is irrelevant). Obviously, the reader can easily add other entities, even extrapolate and correlate the presented parameters with others, for example neural relative power can be partially correlated with an intelligence measurement parameter such as IQ (Intelligence Quotient) like level, etc.

The essential characteristics of neural systems can be identified from the analysis of how nature highlighted them and included in its neural information management systems. Complex informational processes (including intelligent ones) appear in nature with the evolution of living entities towards beings with their own mobility and with other functionalities from the need to manage these functionalities, especially to process the variability of non-deterministic information to which these beings begin to be exposed. Thus, the sensing (afferent) and actuating (efferent) centers are developed, both external (related to sense and locomotors organs, etc.) and internal (related to internal organs, glands, etc.) through automation and creation of specific functions whose essential elements begin to be memorized. Consequently, from the informational perspective, the first two relevant informational characteristics of neural systems are the function and the memory. The management of non-deterministic information was achieved by nature in the neural case similar to the genetic case (see genetic mutations) by using the non-deterministic information itself that it has to process as a required essential processing characteristic, which in the artificial case it can be managed through different methods: stochastic, fuzzy, etc. The nondeterminism is reflected in multiple aspects (degradation, aging, fault tolerance, creativity, etc.) of neural systems from alterations or replacing of smallest informational fragments up to the level of complex processes such as the required unpredictable behavior of a prey to escape from a predator and survive, but also of the predator who must deceive the prey in order not to starve. The effective informational management in neural systems is achieved by the specialization of some cells to transmit information both fragmentally and aggregately to other cells with which they are connected, thus forming neural networks, both in the peripheral area but especially in the central area. The processing of more complex (abstract) information is actually achieved by involving a larger number of processing nodes and connections and alternating the fragmentation and aggregation processes. The basic informational characteristics of a neural system are summarized in Figure 1; many other characteristics are either secondary or reducible to these.

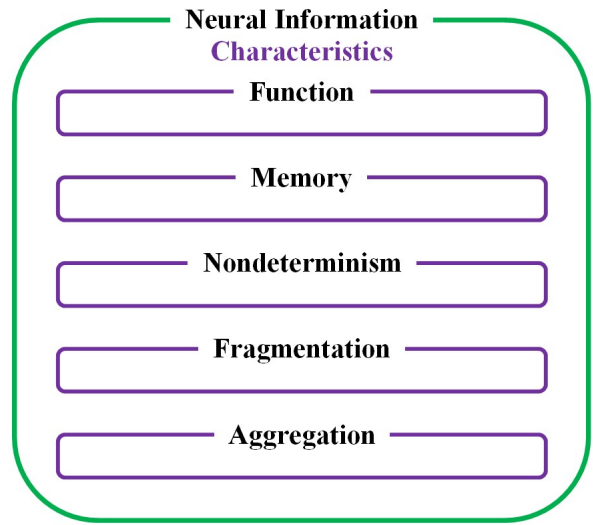

Figure 1. Neural Information Characteristics

The function, as an expression of the functionality, is the main neural characteristic and must be seen starting from the fundamental function of ensuring the survival of the living entity to the specific functions of different organs or internal processes, respectively in performing different tasks or transformations in the artificial case. The memory is the characteristic that completes the functionality of a neural system through the adaptive - contextual storage of relevant information, ensuring to the system: preservation, determinism, coherence, resilience, predictability, etc. The nondeterminism characteristic, as a comprehensive extension of determinism, gives to the neural system solutions to the nondeterministic nature of the environment and system components, ensuring: flexibility, adaptability, originality, creativity, etc. The fragmentation of information by a neural system through neural nodes indicates its cellular (atomic) topology, ensuring the preservation of the informational ensemble even if some fragments are altered. The aggregation is the characteristic of the neural system that encompasses the processes of collection, correlation and integration of informational fragments. These informational features of neural systems are mutually dependent. For example, functions characteristics are: memorable, non-deterministic, fragmenting and constituted by aggregations. Memory is functional, non-deterministic, fragmentary and also constituted by aggregation, and so on.

A neural system can be informationally modeled as a neural machine that should synthesize natural and artificial neural structures and processes, with multiple relevance: biological, scientific, technological, hardware, software, etc. As it appeared in nature, the essence of the neural machine is conceptually related to the function, starting from the fundamental function that is associated with the fundamental objective of the entity, continuing with the particular functions related to some component parts that give them functionality and integration in the system, and ending with simple or very complex internal functions for transforming or abstract some information. The neural machine with all the relevant neural characteristics presented previously is represented in Figure 2.

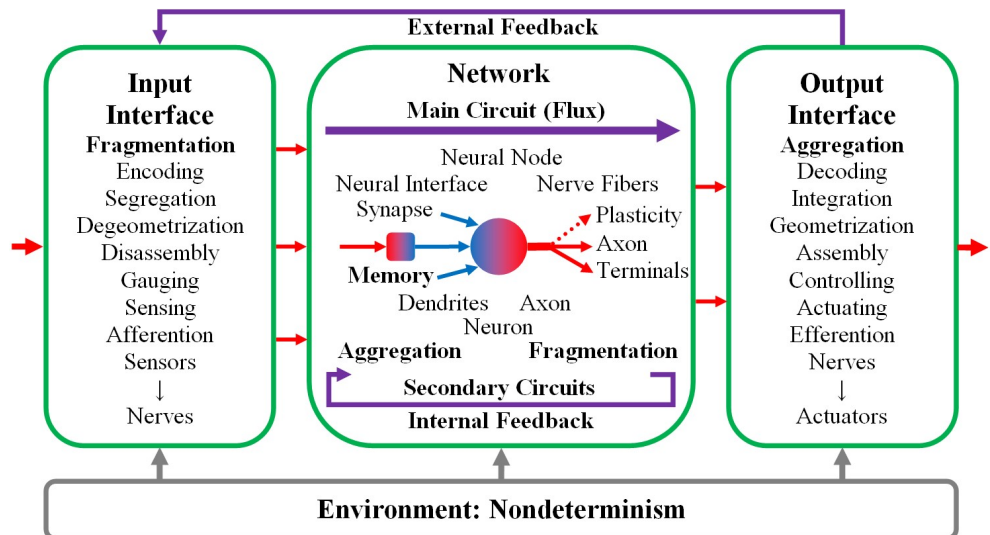

Figure 2. Neural Machine

The information that reaches the input interface is subjected to fragmentation processes, related to transformations such as: encoding, segregation, degeometrization, disassembly, gauging, sensing, etc. The input informational fragments are then processed on a neural network in which the information can go through multiple alternating processes of fragmentation and aggregation. The neural network is a general, non-restrictive, plastic one, with a topology that allows the performing of secondary processes, self connections, internal feedbacks, etc. The information sent by the neural network to the output interface is subjected to aggregation processes such as: decoding, integration, geometrization, assembly, controlling, actuating, efferention, etc. The input and output interfaces can also have a structure similar to that of a neural network, because are reflects the functional essence of neural information, being the compatible source and destination of deeper informational processes in the central area. The neural network itself can be composed by several neural subnetworks or even submachines. A representative neural network example can be seen in Figure 3.

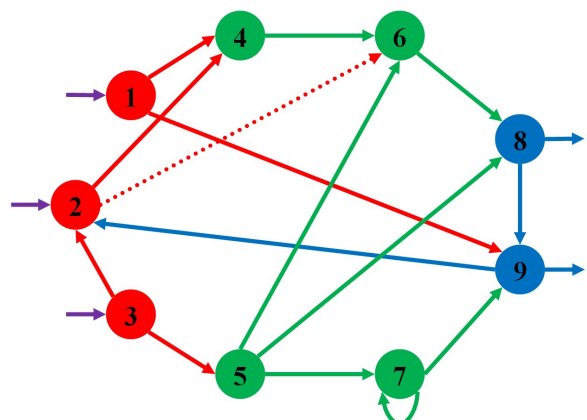

Figure 3. Neural Network Example

The most relevant topological structure that incorporates realistically the neural specificity of information fragmentation and aggregation processes is the generalized (hyper)graph type. In nature, the driving force of phenomena (physical, chemical, biological, etc.) is given by asymmetries and inhomogeneities, whose gradient establishes the semantics of transformations. Idealization through symmetrization, homogenization, stratification, layerization and linearization of non-linear structures and phenomena oversimplifies and generates unrealistic models, with information catastrophes, hallucinations, etc.

A neural system ensures the informational management of the entity's fundamental objective (survival in the natural case, delivering a contextual answer in the case of a ChatBot, etc.) and the related neural machine model can be generally described by

$$\text{Output} = \text{Function}(\text{Input}; \text{History}, \text{Environment}) \quad (3)$$

in which the function is: non-linear, mememorative, nondeterministic, fragmentative, aggregative, etc. The fragmentative-aggregative specificity of neural systems means that both the ensemble and the component parts are modeled in the same way, regardless of scale, the fundamental function of the ensemble being in fact the result of the aggregation of fragments functions (or functionalities) or, in the opposite sense, it can be fragmented into components functions. By explaining the neural characteristics, a functional node or central or peripheral component or the entire neural assembly can be described by:

$$\text{Output} = \text{Neural}(\text{Input}; \text{Function}, \text{Memory}, \text{Nondeterminism}, \text{Fragmentation}, \text{Aggregation}) \quad (4)$$

where the neural characteristics are mutually found in each informational fragment, and by omitting the implicit characteristics, functions associated with representative processes at the level of neural nodes or neural groups (layers, regions, subnets, interfaces, etc.), can be described by:

$$\text{Output} = \text{Aggregation}(\text{Input}), \text{Input} = \text{Fragmentation}(\text{Output}) \quad (5)$$

in which the alternation of fragmentation and aggregation processes offers a more comprehensive perspective compared to the usual approaches as **Aggregation(Input) = Activation(Bias + Weight × Input)** in which the fragments of the input information are (pre)aggregated linearly (through **Bias + Weight × Input**), with the **Activation** function either particularly nonlinear (for example Sigmoid) or only asymmetric (for example ReLU). The linearization of real neural phenomena is a limiting approximation in usual modeling of neural systems, generating a series of informational hallucinations, interferences, inconsistencies, etc. The proposed model incorporates the previous approaches but allows of more realistic description of neural processes, facilitating approximations with superior non-linear terms from the series expansion of the relevant functions (even if they bring them apparently with a cost through a more parameters to manage), because in the biological case, even at the synaptic level, informational fragments are not linear stored and the informational fragments that arrive inside the neurons are not accumulated linearly (additively). Also, the fragmentation-aggregation approach allows the completion of learning methods (such as backpropagation) with continuous training and learning characteristics, as in the natural case.

The present neural information organizing and processing approaches with related neural machines can be easily managed by computational instruments (see Annex 1, 2) being facile incorporable in programming languages libraries or even in language structure [19], in the implementation of formal and natural translators as core methods [17, 18], in both the soft and hard areas as versatile methods, including on related systems such as GPUs, etc. The natural (biological) neural systems have the ability to build flexible internal informational circuits and the proposed neural model facilitates this plasticity and allows the generalization of the circuits used until now in a predilect hardcoded form (architectural, topological, etc.) such as the attention mechanism [16]. Besides facilitating the internal circuits related to some local or temporary informational functions and processes, the model allows the similar neural integration of the input and output interfaces and related circuits as in nervous systems.

Artificial intelligence, viewed as a technological and industrial endeavor to provide smart tools, is a evolutionary rebranding of computers with functions beyond the initial simple mathematical calculations in order to encompass more complex, human-like tasks, incorporating the ability to simulate various aspects of human intelligence, including learning, solving intricate problems, making autonomous decisions, etc. From this perspective, the first computers were actually the first artificial intelligence tools because they were built with the intention of substituting and automating the human task of performing mathematical calculations. Although some methods used in artificial intelligence may not have a direct counterpart in the natural neural systems (some being simple deterministic and conventional numerical methods), neural information methods provide the most relevant deep-level artificial intelligence approach with functions inspired by natural neural processes, but along with this technical and applicative perspective, neural information approach has also the scientific goal to describe the natural neural processes, as relevant as possible.

**Conclusions**

A versatile informational description was presented in order to synthesize the natural neural information specificity and to allow computational neural information relevant implementations in which neural characteristics to be as close as possible to natural ones. In this respect, a series of neural structures, processes, parameters and characteristics were proposed.

In order to describe the informational potential of neural systems regardless of their nature or scale, the general informational parameters as a global quantitative measure of neural systems were introduced through absolute and relative neural power.

The basic neural characteristics that allow a unitary description of neural information regardless the scale were highlighted through: function, memory, nondeterminism, fragmentation and aggregation. These characteristics are found through combinations and interdependence in all components, structures and neural processes, from nodes and connections to regions, networks and interfaces. The processing of more complex information related to more abstract or deep functions is achieved through multiple alternations of information fragmentation and aggregation stages.

The relevant modeling of neural phenomena must take into account their intrinsic non-deterministic characteristics that can be found starting with the informational fragments from the synaptic level to the most complex processes. Even if in practice it can be used in small amounts to increase predictability, in the absence, the neural information only with pure deterministic characteristics will lose their natural essence.

The relevant neural elements were incorporated into a neural machine-type system with input and output interfaces structured similarly to the central neural network based on general graph-type topologies with: inhomogeneities, asymmetries, plasticity, etc. The non-linearity of the processes is also extended to the level of fragmentation and aggregation of information, not only at the node output level through non-linear activations, thus ensuring a more realistic modeling of natural neural processes that are non-linear in depth.

The presented natural and artificial neural information organizing and processing descriptions allow scalable approach of related neuromorphic systems from sensors and actuators information management through simple controllers to nervous systems of robots and versatile neural libraries for realistic artificial intelligence applications.

**References**
[1] W.S. McCulloch, W. Pitts, *A logical calculus of the ideas immanent in nervous activity*, Bulletin of Mathematical Biophysics 5 (1943) 115-133.
[2] D. O. Hebb, *Organization of behavior: A neurophysiological theory*, John Wiley and Sons, 1949.
[3] A. Turing, *Computing Machinery and Intelligence*, Mind 49 (1950) 433-460.
[4] J. Von Neumann, *Probabilistic logics and the synthesis of reliable organisms from unreliable components*, Automata studies 34 (1956) 43-98.
[5] F. Rosenblatt, *The Perceptron: A Perceiving and Recognizing Automaton*, Cornell Aeronautical Laboratory, Report 85-460-1, 1957.
[6] F. Rosenblatt, Frank, *Principles of Neurodynamics: Perceptrons and the Theory of Brain Mechanisms*, Cornell Aeronautical Laboratory, Report VG-1196-G-8, 1962.
[7] J. J. Hopfield, *Neural networks and physical systems with emergent collective computational abilities*, Proceedings of the National Academy of Sciences. 79 (1982) 2554-2558.
[8] D. Rumelhart, G. Hinton, R. Williams, *Learning representations by back-propagating errors*, Nature 323 (1986) 533-536.
[9] B. E. Boser, I. M. Guyon, V. N. Vapnik, *A training algorithm for optimal margin classifiers*, In Proceedings of the Fifth Annual Workshop of Computational Learning Theory 5 (1992) 144-152.
[10] C. Cortes, V. Vapnik, *Support-vector networks*, Machine Learning 20 (1995) 273-297.
[11] S. Hochreiter, J. Schmidhuber, *Long Short-Term Memory*, Neural Computation 9 (1997) 1735-1780.
[12] Y. LeCun, L. Bottou, Y. Bengio, P. Haffner, *Gradient-based learning applied to document recognition*, Proceedings of the IEEE, 86 (1998) 2278-2324.
[13] J. McCarthy, *From here to human-level AI*, Artificial Intelligence 171 (2007) 1174-1182.
[14] A. Krizhevsky, I. Sutskever, G. E. Hinton, *Imagenet classification with deep convolutional neural networks*, Advances in neural information processing systems 25 (2012).
[15] I. Goodfellow, J. Pouget-Abadie, M. Mirza, B. Xu, D. Warde-Farley, S. Ozair, A. Courville, Y. Bengio, *Generative adversarial nets*, Advances in neural information processing systems 27 (2014).
[16] A.Vaswani, N. Shazeer, N. Parmar, J. Uszkoreit, L. Jones, A. N. Gomez, L. Kaiser, I. Polosukhin, *Attention Is All You Need*, Advances in neural information processing systems, 30 (2017).
[17] T. Brown, et al. *Language models are few-shot learners*, Advances in neural information processing systems, 33 (2020) 1877-1901.
[18] I. I. Petrila, *Implementation of general formal translators*, arXiv:2212.08482 (2022).
[19] I. I. Petrila, *@C – augmented version of C programming language*, arXiv:2212.11245 (2022).
[20] G. Novakovsky, N. Dexter, M. W. Libbrecht, W. W. Wasserman, S. Mostafavi, *Obtaining genetics insights from deep learning via explainable artificial intelligence*, Nature Reviews Genetics 24 (2023) 125-137.

[21] P. Pham, L. T. Nguyen, N. T. Nguyen, R. Kozma, B. Vo, *A hierarchical fused fuzzy deep neural network with heterogeneous network embedding for recommendation*, Information Sciences 620 (2023) 105-124.

[22] J. Liu, M. Liu, J.P. Liu, Z. Ye, Y. Wang, Y. Alexeev, J. Eisert, L. Jiang, *Towards provably efficient quantum algorithms for large-scale machine-learning models*, Nature Communications 15 (2024) 434.

**Appendices**

**Annex 1. The simplest C code with the most Neural/AI characteristics**

```c
#include <stdio.h>
#include <stdlib.h>
#include <time.h>

#define Instincts 0//Instinct/Initial/Pre-trained/Start/Birth Information

int Aggregate = Instincts;//Memory/Brain/Network/Parameters/Fragments(Bits)

void Memorization(int Info){Aggregate |= Info;}//Store/Learn

void Association(int Input, int Output){Aggregate |= Input & Output;}//Train

int Transform(int Input){return Aggregate & Input;}//Function/Processing

int Validation(int Input, int Output){return Transform(Input) == Output;}

void Forget(int Info){Aggregate &= ~Info;}//Explicit Forgetting (Erasing)
//..

int main(void)
{
  //srand(time(NULL)); Aggregate |= rand() & rand();//Nondeterminism

  //Memorization(-1);//Saturation/Complete Learning

  Memorization(1); Association(6, 2); Memorization(8);//Learning/Training..

  printf("Identifications: %d, %d\n", Transform(1), Transform(8));//Recognition

  printf("Transformations: 6 => %d\n", Transform(6));//Association/Functionality

  printf("Validation: %d, %d\n", Validation(1, 1), Validation(6, 2));//Confirm

  printf("Determinism: %d\n", Validation(1, 1));//Determinism

  printf("Anticipation: %d\n", Transform(3));//Anticipation/Intuition

  printf("Prediction: 5 => %d\n", Transform(5));//Prediction

  printf("Error: %d instead of 4\n", Transform(4));//Inability/Hallucination

  Aggregate ^= rand() & rand(); printf("Resilience: %d\n", Validation(1, 1));

  //..

  return 0;
}
```

**Annex 2. Neural C Library**

```c
#include <neural.h>

int main(int argc, char* argv[])
{
  Parameter(InputNeurons, 3);

  Parameter(OutputNeurons, 2);

  Parameter(Neurons, 9);//Total Neurons >= InputNeurons + OutputNeurons

  Topology(Connections,//Figure 2 Sample (for explicit topologies)
    0, 0, 0, 1, 0, 0, 0, 0, 1,
    0, 0, 0, 1, 0, 1, 0, 0, 0,
    0, 1, 0, 0, 1, 0, 0, 0, 0,
    0, 0, 0, 0, 0, 1, 0, 0, 0,
    0, 0, 0, 0, 0, 1, 1, 1, 0,
    0, 0, 0, 0, 0, 0, 0, 1, 0,
    0, 0, 0, 0, 0, 0, 1, 0, 1,
    0, 0, 0, 0, 0, 0, 0, 0, 1,
    0, 1, 0, 0, 0, 0, 0, 0, 0);

  //GenerateTopology(Connections, Forward);//Layered, FullGraph etc.
  //Variant for generative topologies, large number of nodes etc.

  NeuralMachine Machine =
  {
    Neurons,//Total Nodes
    Connections,//Topology(Graph)
    InputNeurons,//Inputs Nodes
    OutputNeurons//Outputs Nodes
  };

  SetModel(Machine, BackPropagation);//MonteCarlo etc.

  //ProcessArgs(argc, argv);//Updating some parameters from command line

  Initialize(Machine);

  Train(Machine);

  Test(Machine);

  return 0;
}
```